\documentclass{article}
\usepackage{amsmath,epsfig,graphicx, amsfonts}
\usepackage[preprint]{spconfa4}

\copyrightnotice{978-1-7281-1331-9/20/\$31.00 \copyright 2020 IEEE}

\let\OLDthebibliography\thebibliography
\renewcommand\thebibliography[1]{
  \OLDthebibliography{#1}
  \setlength{\parskip}{0pt}
  \setlength{\itemsep}{0pt plus 0.3ex}
}

\begin{document}\sloppy

\def\x{{\mathbf x}}
\def\L{{\cal L}}

\title{An Edge Information and Mask Shrinking Based Image Inpainting Approach}
%
\name{Huali Xu, Xiangdong Su$^{\ast}$, Meng Wang, Xiang Hao, Guanglai Gao}
\address{Inner Mongolia Key Laboratory of Mongolian Information Processing Technology,\\ College of Computer Science, Inner Mongolia University, Hohhot, China\\ xuhuali.purple@gmail.com, cssxd@imu.edu.cn}

\maketitle

\begin{abstract}
In the image inpainting task, the ability to repair both high-frequency and low-frequency information in the missing regions has a substantial influence on the quality of the restored image. However, existing inpainting methods usually fail to consider both high-frequency and low-frequency information simultaneously. To solve this problem, this paper proposes edge information and mask shrinking based image inpainting approach, which consists of two models. The first model is an edge generation model used to generate complete edge information from the damaged image, and the second model is an image completion model used to fix the missing regions with the generated edge information and the valid contents of the damaged image. The mask shrinking strategy is employed in the image completion model to track the areas to be repaired. The proposed approach is evaluated qualitatively and quantitatively on the dataset Places2. The result shows our approach outperforms state-of-the-art methods.
\end{abstract}
\section{Introduction}
\label{sec:intro}
Image inpainting is the process of reconstructing lost or deteriorated parts of images. The reconstruction quality of both high and low-frequency information in an image dramatically affects the quality of image restoration. The low-frequency information of an image usually more complanate, such as color gradient. The high-frequency information in an image typically includes edges, texture, and other details. Recently, many studies about image inpainting have focused on deep convolutional neural networks\cite{7, 8, 9, 10}, and have achieved remarkable progress. However, these methods ignore the reconstruction of both high-frequency and low-frequency information in the image. They process the missing regions in the same way as the non-missing regions using common convolution. The missing regions contain invalid data. Hence, some invalid information is introduced to the repaired areas in the inpainting process, which leads to an image with an unnatural color transition is generated. Moreover, the generated missing regions usually are over-smoothed and blurry due to the lack of constraints for the high-frequency information reconstruction.

It has been proved that the edge information dramatically affects the ability to reconstruct the high-frequency information and generate fine details for the image. Nazeri et al.\cite{11} proposed an image inpainting model based on a deep convolution neural network. It repairs the missing regions according to edge information, which is generated by an edge generation model. However, this approach ignores the reconstruction of low-frequency information. Compared to the boundary area, it often fails to generate accurate content for the center area in the missing regions, which usually leads to the generated complete image with a weak color convergence and uncomfortable visual artifacts.

Inspired by Liu et al.\cite{12}, we propose the mask shrinking strategy to deal with the issue as mentioned above. This strategy is applied to the image completion model. It includes a special convolution (SConv) as well as a mask updating mechanism. The SConv fills the missing regions using only the valid contents of the damaged image. This means the invalid data of missing areas are discarded in the image inpainting process. Therefore, the missing regions are completed with more visually-realistic content, and the color transition of the repaired image is more natural and realistic. A mask updating mechanism is used after SConv, which records the areas to be restored by shrinking the masked regions.

In summary, the proposed approach restores the damaged regions with edge information as well as the mask shrinking strategy, in which the edge information is beneficial to the reconstruction of high-frequency information while the mask shrinking strategy is good for rebuilding the low-frequency information. We evaluate our model over the public dataset Places2\cite{13} and explain the details of the experiments in Section 5. The result shows that our approach achieves higher-quality inpainting effects. The contributions of this paper are as follows:
\begin{itemize}
\item An adversarial learning approach is proposed to repair both high-frequency information and low-frequency information simultaneously in the image.
\item The complete edge information generated by the edge generation model is good for repairing the high-frequency information of missing regions.
\item The mask shrinking strategy is conducive to restoring the low-frequency information of missing regions.
\end{itemize}

\section{Related Work}
\label{sec:format}
Image inpainting is the process of reconstructing the missing regions of an image. Traditional image inpainting approaches are mainly based on mathematical and physical methods\cite{t2, t3, t5}. In these studies,
Kokaramet et al.\cite{t2} interpolated the defects in a movie with adjacent frames using motion estimation and autoregressive models.
Bertalmio et al.\cite{t5} introduced a new static image restoration algorithm that automatically populates the missing regions with the information surrounding them. However, these approaches usually produce an over-smoothed image and fail to repair excellent results with more details.

Recently, deep learning has been gradually adopted for image inpainting and achieved remarkable progress. Pathak et al.\cite{7} repaired images using an encoder-decoder structure. However, it fails to obtain the fine details of missing regions because of the information bottleneck in the channel-level fully connected layer. Yang et al.\cite{8} proposed a postprocessing approach for context encoders to propagate the texture information from valid regions.
Liu et al.\cite{12} proposed a “partial convolution” to repair the irregular mask holes of images, but this method fails to consider the high-frequency data. Nazeri et al.\cite{11} introduced edge information to constrain the high-frequency information, but they neglect the low-frequency information of images.
The above methods use convolutional neural networks, which do not fully consider both the high-frequency and low-frequency information.

\section{Methods}
\label{sec:format}
\subsection{Network structure}
As shown in Figure 1, the proposed approach contains two models, the edge generation model and the image completion model. They are both GAN based models, which means each of them consists of a generator and a discriminator. The edge generation model is used to generate the complete edge information for the damaged image. $G_{1}$ and $D_{1}$ represent the generator and discriminator of the edge generation model, respectively. $G_{1}$ generates a fake edge map whereas $D_{1}$ aimed to distinguish the fake edge map from the ground truth.

The image completion model repairs the damaged image to a complete image, condition on the complete edge information generated by $G_{1}$. $G_{2}$ and $D_{2}$ respectively express the generator and discriminator of the image completion model. $G_{2}$ generates the fake inpainting result, whereas $D_{2}$ distinguishes the fake inpainting result from the ground truth. The mask shrinking strategy is applied in each layer of $G_{2}$ to reconstruct delicate low-frequency information.
 \begin{figure}[t]

\begin{minipage}[b]{1.0\linewidth}
  \centering
  \centerline{\includegraphics[width=8.7cm]{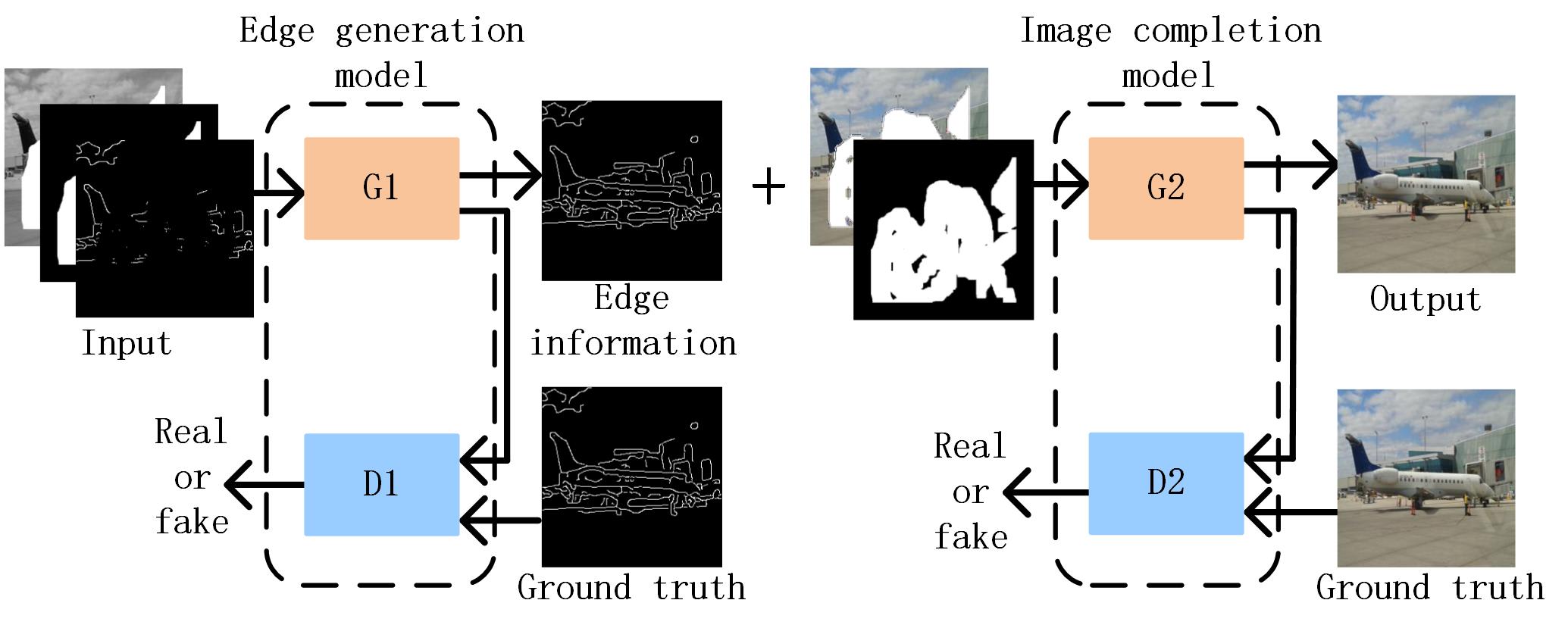}}
\end{minipage}
\caption{Overview of the proposed network structure. It consists of two GAN based models. $G_{1}$ and $D_{1}$ are used to generate the complete edge information, and $G_{2}$ and $D_{2}$ are used to repair the damaged image.}
\label{fig:res}
\end{figure}

We introduce the edge generation model in Section 3.2, illustrate the mask shrinking strategy in Section 3.3, and describe the image completion model in Section 3.4.

\subsection{Edge generation model}
Because of the good performance of the edge generation model in \cite{11}, our edge generation model uses the same structure. The generator $G_{1}$ consists of encoders, residual blocks, and decoders. Encoders down-sample images twice, followed by eight residual blocks, and the decoders then up-sample the feature maps twice to the original image size. Spectral normalization is applied after the convolution in each layer, and dilated convolutions with a dilation factor of two are used in the residual blocks. Discriminator $D_{1}$ uses a $70 \times 70$ PatchGAN\cite{14, 15}. Instance normalization is used across all layers of the network.

\subsection{Mask shrinking strategy}
The mask shrinking strategy is used in each layer of $G_{2}$ to reconstructs the low-frequency information. It includes a special convolution (SConv) followed by a mask updating mechanism. SConv performs a calculation on the damaged image, and the mask updating mechanism updates the mask regions on the mask image. Different from the common convolution process in Figure 2(a), the SConv process in Figure 2(b) penalizes the masked area. This means that SConv repairs the missing regions using only the valid information in the damaged image. SConv can be expressed as follows:
\begin{equation}
\mathbf{S}_{i+1} = \left\{
\begin{aligned}
 &\frac{1}{mean(\mathbf{M})} \mathbf{W}^\mathrm{T}(\mathbf{S}_{i} \odot \mathbf{M}) + \mathit{b}, \quad & mean(\mathbf{M}) > 0 \\
 & 0, & otherwise \\
\end{aligned}
\right.
\end{equation}
where $\mathbf{S}_{i}$ denotes the $i$th layer input pixel values that are consistent with the current convolution window, and $\mathbf{M}$ is the corresponding mask. In addition, $\odot$ denotes element-wise multiplication, and $\mathbf{W}$ and $b$ are the weights and bias for the convolution window, respectively. The term $\frac{1}{mean(\mathbf{M})}$ is used to adjust the effective pixel values.
\begin{figure}[t]

\begin{minipage}[b]{1.0\linewidth}
  \centering
  \centerline{\includegraphics[width=8.5cm]{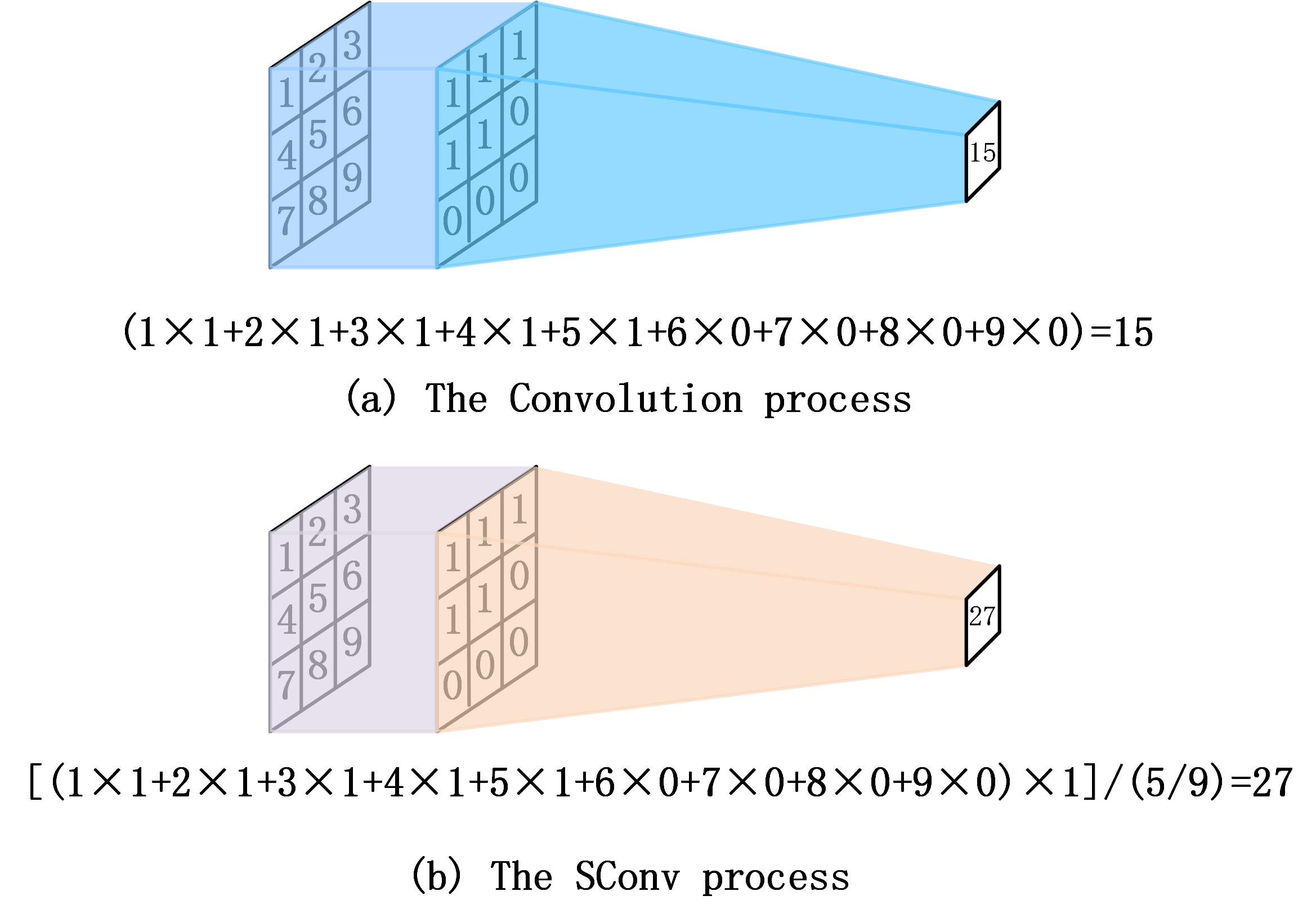}}
\end{minipage}
\caption{Calculation of (a) convolution and (b) SConv}
\label{fig:res}
\end{figure}

The mask updating mechanism shrinks the masked regions of the mask image, which track the repairing status of the damaged image in every layer of $G_{2}$ to guide the repairing of the next layer. It can be considered as another special convolution that can avoid the introduction of noise, in which all the value of the convolution filters is 1. The value would be marked as valid information after the mask updating mechanism when existing least one valid pixel in the current convolution window $\mathbf{M}$. It is expressed as follows:
\begin{equation}
m = \left\{
\begin{aligned}
 &1, \qquad & if \quad mean(\mathbf{M}) > 0 \\
 & 0, & otherwise \\
\end{aligned}
\right.
\end{equation}
where $m$ is a pixel value after the mask updating mechanism: $m=1$ denotes a valid pixel value, and $m=0$ denotes an invalid pixel value.

\subsection{Image completion model}
As shown in Figure 3, the generator $G_{2}$ includes encoders, residual blocks, and decoders. The encoders down-sample the original images four times with modules in the form of convolution-BatchNorm-ReLu\cite{22}, followed by six residual blocks. And the decoders then upsample the feature maps to the original image size with four modules in the form of deconvolution-BatchNorm-LeackyReLu. In the residual blocks, spectral normalization is applied after the convolution in each layer. Each of the convolution in $G_{2}$ is replaced by the mask shrinking strategy. Skip links are used to concatenate the feature maps and masks in the encoders to the feature maps and masks with the same size in the decoders to create the feature map and mask inputs for the next layer of decoders. Similar to $D_{1}$, $D_{2}$ uses a $70\times 70$ patchGAN\cite{14, 15} to distinguish the generated images from the ground truth.
\begin{figure*}[htb]

\begin{minipage}[b]{1.0\linewidth}
  \centering
  \centerline{\includegraphics[width=17cm]{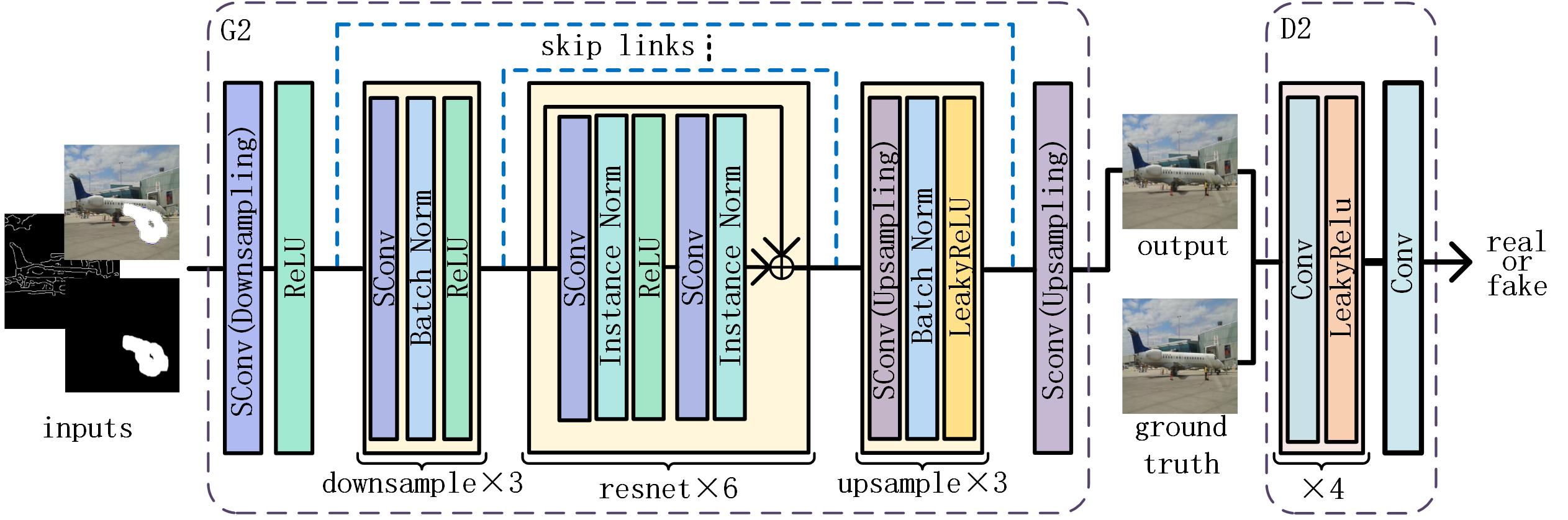}}
\end{minipage}
\caption{The structure of image completion model.}
\label{fig:res}
\end{figure*}

In the image completion network, the damaged image $\widetilde{\textbf{I}}_{gt}$ = $\textbf{I}_{gt} \odot (1 - \textbf{M})$ and the binary mask $\textbf{M}$ (1 for the masked regions) are inputs, which conditional on $\textbf{C}_{comp}$. $\textbf{C}_{comp} = \textbf{C}_{gt} \odot (1 - \textbf{M}) +  \textbf{C}_{pred} \odot \textbf{M}$. $\textbf{C}_{gt}$ is the ground truth of the edge map, and $\textbf{C}_{pred}$ is the complete edge information generated by $G_{1}$. The output of the image completion network is a complete image, which has the same size as the input image.
\begin{equation}
\begin{aligned}
\textbf{I}_{pred} = \emph{G}_{2}(\widetilde{\textbf{I}}_{gt}, \widetilde{\textbf{C}}_{comp}, \textbf{M})
\end{aligned}
\end{equation}

The loss function consists of an $\emph{l}_{1}$ loss, an adversarial loss, a perceptual loss, and a style loss. The adversarial loss is
\begin{equation}
\begin{aligned}
L_{adv,2} = \mathbb{E}_{(\textbf{I}_{gt}, \textbf{C}_{comp})}[log \emph{D}_{2}(\textbf{I}_{gt}, \textbf{C}_{comp})]\\
+\mathbb{E}_{\textbf{C}_{comp}}log[1 - \emph{D}_{2}(\textbf{I}_{pred}, \textbf{C}_{comp})]
\end{aligned}
\end{equation}

Perceptual loss $L_{prec}$ constrains the similarity to ground truth by calculating the distance between activation maps. It is defined as
\begin{equation}
\begin{aligned}
L_{prec} = \mathbb{E} \left[  \sum_{i} \cfrac{1}{N_{i}} \left\|  \phi_{i}(\textbf{I}_{gt}) - \phi_{i}(\textbf{I}_{pred}) \right \|_{1}  \right]
\end{aligned}
\end{equation}
where $\phi_{i}$ is the activation map of the \emph{i}’th layer of the network.

Assuming that the size of feature maps is $C_{j} \times H_{j} \times W_{j}$, style loss $L_{style}$ is expressed as follows:
\begin{equation}
\begin{aligned}
L_{style} = \mathbb{E}_{j} \left[ \left\|    G_{j}^{\phi}(\widetilde{\textbf{I}}_{pred}) - G_{j}^{\phi}(\widetilde{\textbf{I}}_{gt})    \right \|_{1}  \right]
\end{aligned}
\end{equation}
where $G_{j}^{\phi}$ is a $C_{j} \times C_{j}$ Gram matrix constructed from activation maps $\phi_{j}$. In summary,the objective funciton is
\begin{equation}
\begin{aligned}
L_{G_{2}} = \lambda_{l_{1}}L_{l_{1}} + \lambda_{adv,2}L_{adv,2} + \lambda_{p}L_{prec} + \lambda_{s}L_{style}
\end{aligned}
\end{equation}

We set the parameters in the loss function as follows: $\lambda_{l_{1}}$ =1, $\lambda_{adv,2}$ = $\lambda_{p}$ = 0.1, $\lambda_{s}$ = 250.

\section{Experiments}
\label{sec:pagestyle}
\subsection{Datasets}
Our datasets include an image dataset, a mask dataset, and an edge maps dataset. For the image dataset, we choose 5,000 images in Places2, in which 4,000 images were used for training and 1,000 images were used for testing. The mask dataset is from Liu et al.\cite{11}. We employ 20,000 irregular mask images, in which 19,000 mask images are used for the training phase and the remaining 1,000 mask images are used in the testing phase. The edge ground truth is generated by the Canny edge detector\cite{17}. We set the sensitivity of the Canny edge detector as 2, which controlled by the standard deviation of the Gaussian smoothing filter.
\subsection{Metrics}
The metrics include peak signal-to-noise ratio (PSNR), structural similarity index (SSIM), and the Frechet inception distance (FID)\cite{FID}. The PSNR and SSIM indices are two indicators for evaluating the similarity of images, which are widely used in the field of image processing. The FID is commonly used to evaluate the results of a generative adversarial network (GAN). It is expressed as
\begin{equation}
\begin{aligned}
FID(\mathbb{P}_{r}, \mathbb{P}_{g})=|| \mu_{r}-\mu_{g} || + Tr(\textbf{C}_{r}+\textbf{C}_{g}-2(\textbf{C}_{r}\textbf{C}_{g})^{1/2})
\end{aligned}
\end{equation}
The FID also reflects the quality of the images that are generated by GAN.
\section{Results}
The results of a qualitative and quantitative comparison for the proposed approach with current state-of-the-art methods are given in this section. The results show that the proposed approach obtains the best effects of qualitatively and quantitatively. We denote the common convolution as $C$ and represent the convolution using mask shrinking strategy as $CM$. $n(CM)-i(C)-j(CM)$ means the structure consists of $n$ downsampling with $CM$ followed by $i$ residual blocks with $C$ and $j$ upsampling with $CM$.
\subsection{Selection of best network structure}
To determine the optimal network structure of the image completion model, we compare the performance of different structures. The results are listed in Table \ref{tab1}. 
As $4(CM)-6(CM)-4(CM)$ structure obtains the best score for all metrics as shown in Table \ref{tab1}, we use it as the default structure of our model in the following experiments.
\begin{table}[t]
\centering
\caption{Evaluation of different structures} \label{tab1}
\scalebox{1}{
\begin{tabular}{|l|c|c|c|c|}
  \hline
  model           & PSNR        & SSIM           &FID   \\  \hline  
  $5(C)-4(C)-5(C)$           &32.44  &0.702           &9.58       \\   
  $6(C)-2(C)-6(C)$           &32.20  &0.692          &9.58       \\   
  $4(C)-6(C)-4(C)$             &31.96      &0.696  &9.69  \\ 
  
  $7(C)-0(C)-7(C)$ \cite{21}            & 31.41   &0.578           & 11.69       \\   
  
  $5(CM)-4(C)-5(CM)$           & 32.56 &0.712           &8.84       \\   
  $6(CM)-2(C)-6(CM)$           &32.11  &0.710           &8.22       \\   
  $4(CM)-6(C)-4(CM)$             &32.46  &0.711  &8.99   \\ 
  
  $7(CM)-7(CM)$             &31.82          &0.641         &10.90        \\   
  $5(CM)-4(CM)-5(CM)$           &32.49  &0.713           &8.40        \\   
  $6(CM)-2(CM)-6(CM)$           &32.56 &0.713         &8.50        \\   
  $4(CM)-6(CM)-4(CM)$             &\textbf{32.74}          & \textbf{0.717} & ~~\textbf{8.12}  \\  \hline 
\end{tabular}}
\end{table}
\subsection{Comparision with baselines}
Table \ref{tab2} compares the results of the proposed model with those of the current state-of-the-art approaches.
Our approach obtains the best results with a PSNR of 32.74, SSIM of 0.717, and FID of 8.12. Hence, our approach that combines the edge information with the mask shrinking strategy is more effective than other image inpainting approaches.
\begin{table}[t]
\centering
\caption{Evaluation of different approaches} \label{tab2}
\scalebox{1}{
\begin{tabular}{|l|c|c|c|c|}
  \hline
  model           & ~PSNR~        & ~SSIM~           & ~FID~~   \\  \hline  
  PConv\cite{12}           & ~31.02~  &~0.553~           &~13.32~~       \\   
  Pluralistic Inpainting\cite{Plu} &~32.47~  &~0.665~          &~9.40~~      \\   
  EdgeConnect\cite{11}            &~32.55~      &~0.710~  &~9.64~~ \\ 
  Our apprroach            &~\textbf{32.74}~          & ~\textbf{0.717}~ & ~\textbf{8.12}~~  \\  \hline 
\end{tabular}}
\end{table}

Figure 4 displays the inpainting results of different approaches. It is clearly that the images generated by our approach are closer to the ground truth than that generated by other methods.
\begin{figure*}[t]

\begin{minipage}[b]{1.0\linewidth}
  \centering
  \centerline{\includegraphics[width=16.5cm]{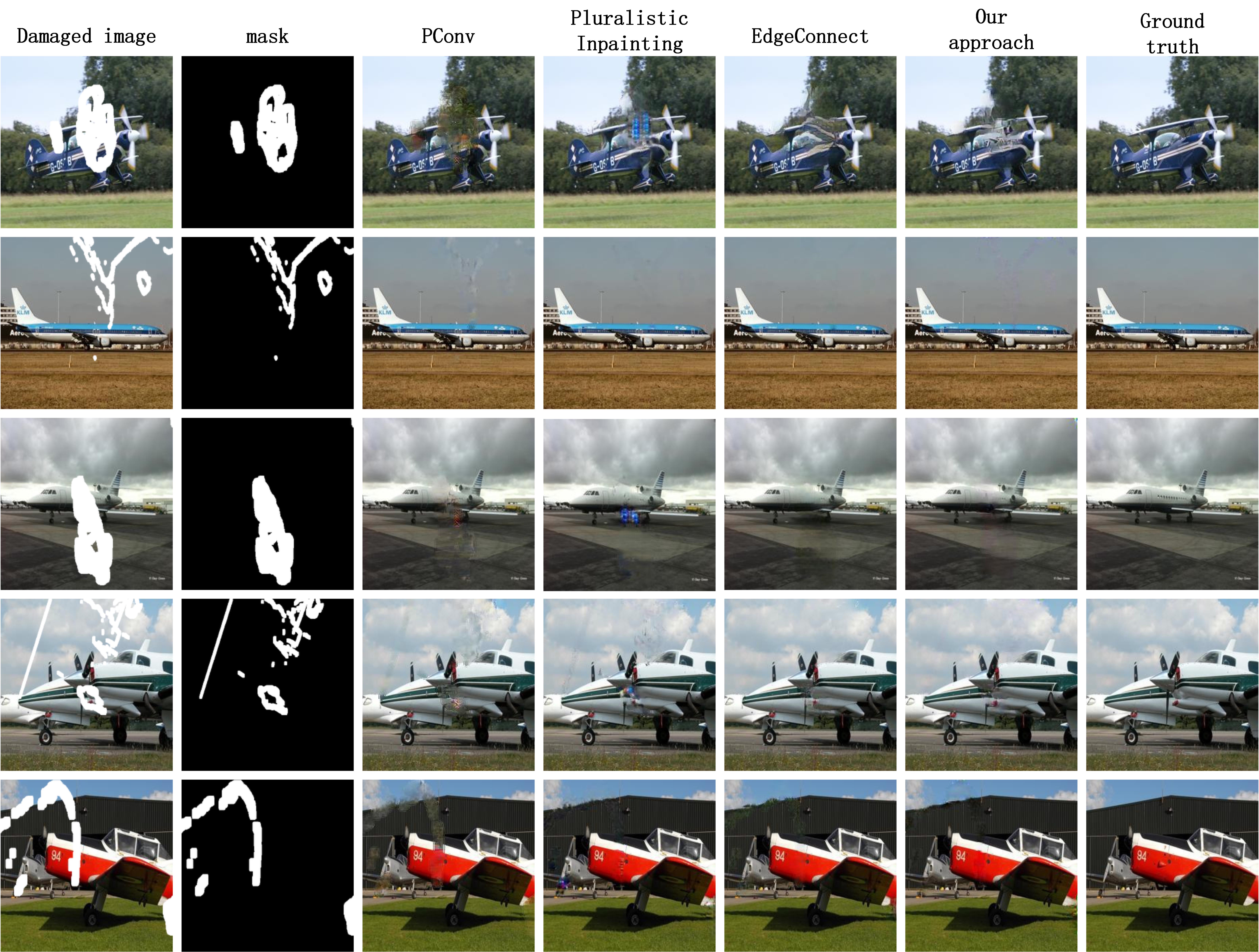}}
\end{minipage}
\caption{Image results of different methods}
\label{fig:res}
\end{figure*}
\subsection{Effect of edge information}
We evaluate the impact of edge information. The results list in Table \ref{tab3}, in which $Yes$ and $No$ represent the results with and without edge information, respectively. The result shows the former obtains more competitive effects than the latter. The results with edge information obtain more competitive effect than that without edge information, indicating that joining the edge information can improve the quality of the repaired image. Hence, the edge information has a strong impact on the quality of the repaired image.

\begin{table}[t]
\centering
\caption{The impact of edge information for the inpainting result.} \label{tab3}
\scalebox{1.1}{
\begin{tabular}{|l|c|c|c|c|}
  \hline
  edge information     & ~~PSNR~~        & ~~SSIM~~           &~~FID~~   \\  \hline  
  $No$    &32.13  &0.677      &9.35       \\   
  $Yes$      &\textbf{32.74}         &\textbf{0.717}        &\textbf{8.12}        \\ \hline  
\end{tabular}}
\end{table}
\subsection{Effect of mask shrinking strategy}
To verify the validity of the mask shrinking strategy, we compare the performance resulting from three different structures, $4(C)-6(C)-4(C)$, $4(CM)-6(C)-4(CM)$ and $4(CM)-6(CM)-4(CM)$.
As shown in Table \ref{tab4}, the result shows that $4(CM)-6(CM)-4(CM)$ obtains best results.
\begin{table}[t]
\centering
\caption{The impact of the mask shrinking strategy for the inpainting result.} \label{tab4}
\scalebox{1.0}{
\begin{tabular}{|l|c|c|c|}
  \hline
  model & PSNR & SSIM & FID \\
  \hline
  $4(C)-6(C)-4(C)$          & 31.96              & 0.696 & 9.69 \\
  $4(CM)-6(C)-4(CM)$    & 32.46              & 0.711 & 8.99 \\
  $4(CM)-6(CM)-4(CM)$ & \textbf{32.74}  & \textbf{0.717} & \textbf{8.12}  \\
  \hline
\end{tabular}}
\end{table}
\subsection{Effect of skip links}
Skip links connect the downsampling and upsampling in our network. We evaluate the effect in Table \ref{tab5}. $Yes$ and $No$ represent the network with and without skip links, respectively. The result shows that overall repair performance is improved by using skip links.

\begin{table}[t]
\centering
\caption{The impact of skip links for the inpainting result} \label{tab5}
\scalebox{1.1}{
\begin{tabular}{|l|c|c|c|c|c|c|}
  \hline
  skip links~~~~ & ~~~PSNR~~~ & ~~~SSIM~~~ & ~~~FID~~~ \\
  \hline
  $No$ & 32.71  & 0.707  & 8.85   \\
  $Yes$ &\textbf{32.74} & \textbf{0.717} & \textbf{8.12}  \\
  \hline
\end{tabular}}
\end{table}
\section{Conclusion}
\label{sec:typestyle}
This paper proposes an edge information and mask shrinking based image inpainting approach. It consists of two GAN based models, i.e., an edge generation model and an image completion model. The edge generation model repairs the edge information for the damaged image. The edge information is used in the image completion model, which repairs the damaged image to a complete image. In the image completion model, we introduce a mask shrinking strategy that repairs images with a special convolution(SConv) and tracks the repairing process with a mask updating mechanism. Experiments on the public dataset Places2 and a comparison with current state-of-the-art methods demonstrate that the proposed approach achieves the best performance.

\section{Acknowledgements}
This work was funded by National Natural Science Foundation of China (Grant No. 61762069, 61773224), Natural Science Foundation of Inner Mongolia Autonomous Region (Grant No. 2019ZD14, 2017BS0601) and Science and technology program of Inner Mongolia Autonomous Region (2019).  

\bibliographystyle{IEEEbib}
\bibliography{icme2020template}

\begin{thebibliography}{10}

\bibitem{7}
Deepak Pathak, Philipp Krahenbuhl, Jeff Donahue, Trevor Darrell, and Alexei~A.
  Efros,
\newblock ``Context encoders: Feature learning by inpainting,''
\newblock in {\em IEEE Conference on Computer Vision \& Pattern Recognition},
  2016.

\bibitem{8}
Yang Chao, Lu~Xin, Lin Zhe, Eli Shechtman, Oliver Wang, and Li~Hao,
\newblock ``High-resolution image inpainting using multi-scale neural patch
  synthesis,''
\newblock 2016.

\bibitem{9}
Ruohan Gao and Kristen Grauman,
\newblock ``On-demand learning for deep image restoration,''
\newblock 2017.

\bibitem{10}
Satoshi Iizuka, Edgar Simo-Serra, and Hiroshi Ishikawa,
\newblock {\em Globally and locally consistent image completion},
\newblock 2017.

\bibitem{11}
Kamyar Nazeri, Eric Ng, Tony Joseph, Faisal Qureshi, and Mehran Ebrahimi,
\newblock ``Edgeconnect: Generative image inpainting with adversarial edge
  learning,''
\newblock 2019.

\bibitem{12}
Guilin Liu, Fitsum~A. Reda, Kevin~J. Shih, Ting{-}Chun Wang, Andrew Tao, and
  Bryan Catanzaro,
\newblock ``Image inpainting for irregular holes using partial convolutions,''
\newblock {\em CoRR}, vol. abs/1804.07723, 2018.

\bibitem{13}
B.~Zhou, A~Lapedriza, A~Khosla, A~Oliva, and A~Torralba,
\newblock ``Places: A 10 million image database for scene recognition,''
\newblock {\em IEEE Trans Pattern Anal Mach Intell}, vol. PP, no. 99, pp. 1--1,
  2018.

\bibitem{t1}
Mark Nitzberg, David Mumford, and Takahiro Shiota,
\newblock {\em Filtering, segmentation and depth}, vol. 662,
\newblock Springer, 1993.

\bibitem{t2}
Anil~C Kokaram, Robin~D Morris, William~J Fitzgerald, and Peter~JW Rayner,
\newblock ``Interpolation of missing data in image sequences,''
\newblock {\em IEEE Transactions on Image Processing}, vol. 4, no. 11, pp.
  1509--1519, 1995.

\bibitem{t3}
Anil~N Hirani and Takashi Totsuka,
\newblock ``Combining frequency and spatial domain information for fast
  interactive image noise removal,''
\newblock in {\em SIGGRAPH}, 1996, vol.~96, pp. 269--276.

\bibitem{t4}
Simon Masnou and J-M Morel,
\newblock ``Level lines based disocclusion,''
\newblock in {\em Proceedings 1998 International Conference on Image
  Processing. ICIP98 (Cat. No. 98CB36269)}. IEEE, 1998, pp. 259--263.

\bibitem{t5}
Marcelo Bertalmio, Guillermo Sapiro, Vincent Caselles, and Coloma Ballester,
\newblock ``Image inpainting,''
\newblock in {\em Proceedings of the 27th annual conference on Computer
  graphics and interactive techniques}. ACM Press/Addison-Wesley Publishing
  Co., 2000, pp. 417--424.

\bibitem{21}
Olaf Ronneberger, Philipp Fischer, and Thomas Brox,
\newblock ``U-net: Convolutional networks for biomedical image segmentation,''
\newblock in {\em International Conference on Medical image computing and
  computer-assisted intervention}. Springer, 2015, pp. 234--241.

\bibitem{22}
Sergey Ioffe and Christian Szegedy,
\newblock ``Batch normalization: Accelerating deep network training by reducing
  internal covariate shift,''
\newblock {\em arXiv preprint arXiv:1502.03167}, 2015.

\bibitem{14}
Phillip Isola, Jun-Yan Zhu, Tinghui Zhou, and Alexei~A Efros,
\newblock ``Image-to-image translation with conditional adversarial networks,''
\newblock in {\em Proceedings of the IEEE conference on computer vision and
  pattern recognition}, 2017, pp. 1125--1134.

\bibitem{15}
Jun-Yan Zhu, Taesung Park, Phillip Isola, and Alexei~A Efros,
\newblock ``Unpaired image-to-image translation using cycle-consistent
  adversarial networks,''
\newblock in {\em Proceedings of the IEEE international conference on computer
  vision}, 2017, pp. 2223--2232.

\bibitem{17}
J~Canny,
\newblock ``A computational approach to edge detection,''
\newblock {\em IEEE Trans. Pattern Anal. Mach. Intell.}, vol. 8, no. 6, pp.
  679--698, June 1986.

\bibitem{IS1}
Christian Szegedy, Vincent Vanhoucke, Sergey Ioffe, Jon Shlens, and Zbigniew
  Wojna,
\newblock ``Rethinking the inception architecture for computer vision,''
\newblock in {\em Proceedings of the IEEE conference on computer vision and
  pattern recognition}, 2016, pp. 2818--2826.

\bibitem{IS2}
Tim Salimans, Ian Goodfellow, Wojciech Zaremba, Vicki Cheung, Alec Radford, and
  Xi~Chen,
\newblock ``Improved techniques for training gans,''
\newblock in {\em Advances in neural information processing systems}, 2016, pp.
  2234--2242.

\bibitem{FID}
Martin Heusel, Hubert Ramsauer, Thomas Unterthiner, Bernhard Nessler, and Sepp
  Hochreiter,
\newblock ``Gans trained by a two time-scale update rule converge to a local
  nash equilibrium,''
\newblock in {\em Advances in Neural Information Processing Systems}, 2017, pp.
  6626--6637.

\end{thebibliography}

\end{document}